\pdfoutput=1

\documentclass[11pt]{article}

\usepackage[]{acl}

\usepackage{times}
\usepackage{latexsym}

\usepackage{algpseudocode}
\usepackage{graphicx}
\usepackage{booktabs}

\usepackage[T1]{fontenc}

\usepackage[utf8]{inputenc} 

\newcommand{\yz}[1]{}
\renewcommand{\yz}[1]{\textcolor{blue}{\textbf{Zhu}: #1}}

\title{SiRA: Sparse Mixture of Low Rank Adaptation}

\author{
Yun Zhu\thanks{Correspondence to yunzhu@google.com .} \qquad
Nevan Wichers\footnotemark[1] \qquad 
Chu-Cheng Lin\footnotemark[1] \qquad 
Xinyi Wang\footnotemark[1] \qquad
Tianlong Chen\footnotemark[2] \qquad
\AND
Lei Shu\footnotemark[1] \qquad
Han Lu\footnotemark[1] \qquad
Canoee Liu\footnotemark[1] \qquad
Liangchen Luo\footnotemark[1] \qquad
Jindong Chen\footnotemark[1] \qquad
Lei Meng\footnotemark[1] \\ \\
Google\footnotemark[1],  CSAIL@MIT\footnotemark[2]
}

\begin{document}
\maketitle
\begin{abstract}
Parameter Efficient Tuning has been an prominent approach to adapt the Large Language Model to downstream tasks. 
Most previous works considers adding the dense trainable parameters, where all parameters are used to adapt certain task. 
We found this less effective empirically using the example of LoRA that introducing more trainable parameters does not help.
Motivated by this we investigate the importance of leveraging ``sparse'' computation and propose
SiRA: sparse mixture of low rank adaption. 
SiRA leverages the Sparse Mixture of Expert(SMoE) to boost the performance of LoRA. 
Specifically it enforces the top $k$ experts routing with a capacity limit restricting the maximum number of tokens each expert can process. 
We propose a novel and simple expert dropout on top of gating network to reduce the over-fitting issue. 
Through extensive experiments, we verify SiRA performs better than LoRA and other mixture of expert approaches across different single tasks and multitask settings. 


\end{abstract}

\section{Introduction}

Large Language Models~\citep{thoppilan2022lamda, palm2, ouyang2022training,touvron2023llama} (LLMs) have demonstrated impressive capabilities in a wide range of tasks.
Yet to adapt these general-purpose models to downstream low resource tasks is especially important.
To this end parameter efficient tuning (PET)~\citep{hu2021lora, li2021prefix, lester2021power, houlsby2019parameter, zhang2023llama, zaken2021bitfit, chen2022revisiting}, which introduces task specific weights to the frozen foundation model for gradient descent, avoids catastrophic forgetting~\citep{luo2023empirical} of fine-tuning and offers better quality and lower cost than in-context learning~\citep{liu2022few}. 
LoRA~\citep{hu2021lora} is a widely adopted PET method which achieves high performance leveraging low-rank matrices. 



One question for user of LoRA or other PET approach is how much trainable parameter should be used.
In the case of LoRA, it is controlled by the rank of the low-rank matrices. And by increasing this value more computation could be provided to fit specific tasks.
However it has been shown higher rank matrices will not introduce better quality to the model due to the instability of training~\citep{chen2022revisiting}, which we verify in Figure~\ref{fig:comp} in Appendix \ref{app:comp}.
This poses a hidden bottleneck for the quality of the model even when we have enough computation budget, and it remains a challenge to improve the quality of LoRA. 


On the other hand, leveraging Sparse Mixture of Experts in neural networks has been extensively studied as a replacement for FeedForward Networks, with different approaches to find the optimal assignment between experts and tokens, including reinforcement learning~\citep{bengio2015conditional}, linear programs~\citep{lewis2021base}, fixed rules~\citep{roller2021hash}, top-1 gating~\citep{fedus2022switch}, top-2 gating~\citep{lepikhin2020gshard}, top-k gating~\citep{shazeer2017outrageously}, reverse expert choosing~\cite{zhou2022mixture}, and soft assignment~\cite{puigcerver2023sparse}. 
Several recent works have proposed to utilize mixture-of-expert models on top of parameter-efficient tuning~\citep{wang2022adamix,zadouri2023pushing}. However, prior works overlooked the potential of sparse MoE (SMoE), arguably because of issues like token dropping~\citep{puigcerver2023sparse} and overfitting~\citep{elbayad2022fixing}.



To this end, we investigate leveraging ``sparse'' computation for PET in this paper and propose \textbf{SiRA}, the Sparse Mixture of Low Rank Adaptation. 
SiRA enforces the top $k$ experts routing, which improves resource and computation utilization and also facilitates more fine-grained capacity control of a given input. 
SiRA consists of three important ingredients: the capacity constraint which embraces token dropping, an auxiliary loss to penalize over-utilizing few experts, and a novel expert dropout mechanism. They work together to ensure the proper load balancing and address the over-fitting issue.






We conducted extensive experiments to verify the performance of SiRA, which achieves better performance than LoRA~\citep{hu2021lora} and its MoE variants Adamix~\citep{wang2022adamix}, MoLoRA~\citep{zadouri2023pushing} across wide range of single task and multitask benchmarks. 
Our ablation study showed that the number of used experts and capacity per expert improves performance which demonstrates advantage of being ``sparse''. 
Notably the expert dropout plays an important role, and it is more effective than SMoE-dropout~\citep{chen2023sparse}.

\section{Related Works}

\paragraph{Parameter Efficient Tuning (PET)}

Parameter Efficient Tuning has a variety of flavors such as Adapters~\citep{houlsby2019parameter}, Prefix Tuning~\citep{li2021prefix, liu2021p}, Prompt Tuning~\citep{lester2021power}, P-tuning~\citep{liu2023gpt}, attention-injection~\citep{zhang2023llama}, LoRA~\citep{hu2021lora, dettmers2023qlora}, and combinations of PET methods~\citep{mao2021unipelt}. 
In this paper, our focus on LoRA as it has been found to achieve better results, although the methods could be applied to other flavors as well.



\paragraph{MoE for PET methods}
Along the intersection of PET and MoE, Adamix~\citep{wang2022adamix} and MoLoRA~\cite{zadouri2023pushing} are most similar to our work. Adamix randomly chooses an expert in training and averages all the experts during inference. Albeit efficient, this method is more like checkpoint averaging~\citep{gao2022revisiting} because the experts are randomly chosen they don't learn to specialize. More importantly, the random approach has significantly longer training time, which is multiplied by the number of experts used. 
MoLoRA applies the full soft MoE on the top of LoRA, where all experts are averaged using a learned gating. 
Compared to this work, our method can achieve better efficiency. Firstly, SiRA does not need longer time to train compared to standard LoRA, thanks to the quick convergence of SMoE~\citep{fedus2022switch}. Secondly, the sparsity is enforced in SiRA which saves the training resource and inference computation compared to MoLoRA. 

Another track of the MoE work is for multitasking, such as Task-MoE~\citep{kudugunta2021beyond} and Skill Selection~\citep{ponti2023combining}. These approaches assume the external task-id as an extra input for training and inference. Although we experiment with MoE in multitask settings, it does not require the task-id of inputs. 


\section{Sparse Mixture of Low Rank Adaptation}

To increase the capacity of LoRA~\citep{hu2021lora} using Mixture of Experts~(MoE) without adding too much computational cost, we propose Sparse Mixture of Experts of Low Rank Adaptation~(SiRA), which leverages multiple lightweight LoRA adaptors as experts while enforcing sparsity when using the experts. 

\begin{figure}[!t]
\includegraphics[width=0.98\columnwidth]{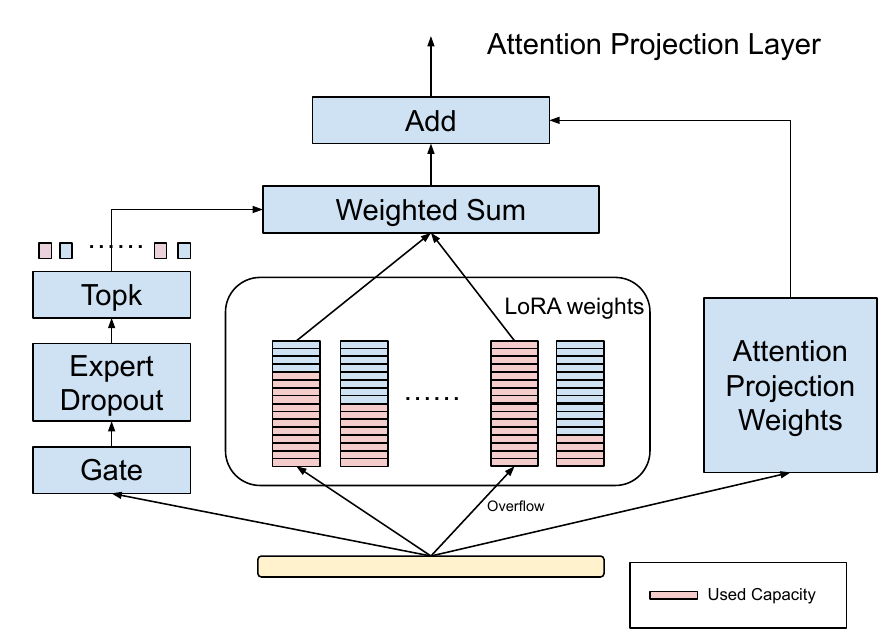}
\centering
\caption{SiRA: Sparse Gated Mixture of LoRA.
}
\label{fig:sira}
\end{figure}

Figure~\ref{fig:sira}~shows an illustration of SiRA. The MoE layer for the adapter consists of $E$ experts, each with their own LoRA weights, $W_1, . . ., W_E$. $W_k$ is the product of two low rank matrices $W_k = B_k A_k$. 
We also assume the base foundation model has $W_0$ as it is frozen weight, which represents either query, key, value, or output projection.  
Next, we will introduce each component of our framework that enforces the sparsity of LoRA experts.


\paragraph{Expert Gating}
To reduce the computational cost, SiRA only actives a subset of all the expert modules. Formally, during each forward pass, we select $K$ out of $E$ experts using the output scores of a gating network  $\theta_g$. 
The process is mathematically expressed as Equation~(\ref{eq:topk}) and (\ref{eq:sum}), where $s$ denote the token index of the sequence $x$ and $G_{s, e}$ is the gating network output at $s$-th token $e$-th experts.  

\begin{equation}
\label{eq:topk}
    G(x_s) =  \mathrm{TopK}(\mathrm{softmax}(\theta_g^T x_s))
\end{equation}

\begin{equation}
\label{eq:sum}
    y_s = \sum_{e=1}^{E} G_{s, e} W_e(x_s) + W_0(x_s)
\end{equation}

\paragraph{Experts Dropout}
We propose a practical way to make the gate more balanced with a simple gate dropout. Specifically, we introduce dropout to the gating output $G$ as shown in Equation~\ref{eq:dropout}.

\begin{equation}
\label{eq:dropout}
    G(x_s) = \mathrm{TopK}(\mathrm{Dropout}(\mathrm{softmax}(\theta_g^T x_s)))
\end{equation}


\paragraph{Expert Token Capacity}
We enforce the capacity constraints for experts following GShard~\citep{lepikhin2020gshard}. 
Specifically, we restrict that the number of tokens processed by each expert should not exceed a predefined threshold. 
Once the capacity is reached, one expert simply drop the overflow tokens. 
If all $K$ experts reach their token capacity before all tokens in a training example is processed, the rest of the tokens will only be encoded using the frozen model parameter $W_0$.

\paragraph{Auxiliary Loss}
We define an auxiliary loss term to further encourage load balancing among different experts~\citep{shazeer2017outrageously}. We define the total number of tokens to be $S$, and there is $E$ experts. We denote $c_e$ as the number of tokens routed to expert $e$.
By using the mean gates per expert $m_e=Mean_{s}(\mathrm{Dropout}(\mathrm{softmax}(\theta_g^T x_s)))$ as a differentiable approximation, 
we define the aux loss in Equation~\ref{eq:aux}.

\begin{equation}
    l_{aux} = \frac{1}{E} \sum_{e=1}^E \frac{c_e}{S}*m_e
    \label{eq:aux}
\end{equation}


\section{Experiments}

\begin{table*}[t]
\caption{Performance Comparison For Single Tasks}
\fontsize{8.2pt}{8.2pt}\selectfont
\setlength{\tabcolsep}{2pt}
\begin{tabular}{l|c|cc|ccc|c|c|c|c|c|c}
\toprule
 Approach & $\delta$ Params &
  \multicolumn{2}{c}{FinQA (EN)} &
  \multicolumn{3}{c}{ForumSum (EN)} &
  \multicolumn{1}{c}{SP (SW)} &
  \multicolumn{1}{c}{QA-in (SW)} &
  \multicolumn{1}{c}{NER (SW)} &
  \multicolumn{1}{c}{SP (BN)} &
  \multicolumn{1}{c}{QA-in (BN)} &
  \multicolumn{1}{c}{QA-cross (BN)}
  \\ \midrule
         &   & em     & f1     & bleurt     & rougeL    & f1     & accuracy     & f1     & span-f1     & accuracy     & f1 & f1    \\     \midrule                       
LoRA & 0.043\% & $5.0$ & $5.6$ & $96.70$ & $33.97$ & $23.54$ & $27.63$ & $82.08$ & $88.95$ & $33.52$ & $80.34$ & 76.81 \\




Adamix & 0.689\% &5.6 & 6.0 & 95.95 & 35.10 & 23.88 & \textbf{33.22}  & 81.24 & 89.00 & \textbf{39.03} & 81.70 & 76.07 \\

MoLoRA & 0.746\% & 5.6& 6.4 & 97.05 & 34.37 & 24.79 &  32.50 & 82.33 & 89.33 & 36.28 & 79.06  & 76.75  \\

SiRA & 0.746\% & \textbf{5.8} & \textbf{6.6} & \textbf{97.14} & \textbf{35.67} & \textbf{25.83} & 32.52 & \textbf{83.00} & \textbf{89.95} & 38.61 & \textbf{82.10} & \textbf{76.93} \\

\bottomrule
\end{tabular}
\label{tab:single_task}
\end{table*}


\begin{table*}[t]
\caption{Performance Comparison For Multi Tasks}
\fontsize{9pt}{9pt}\selectfont
\setlength{\tabcolsep}{2pt}
\begin{tabular}{l|c|cccc|cccc}
\toprule
 
 Approach & $\delta$ params &
  \multicolumn{4}{c}{SW Multitask} &
  \multicolumn{4}{c}{BN Multitask} 
  \\ \midrule
           & & SP(accuracy)     & QA-in(f1)     & NER(span-f1)   &  Average  & SP(accuracy)     & QA-in(f1) & QA-cross(f1)   &  Average \\     \midrule                       
LoRA &0.043\% &28.06 & 77.71 & 88.28 & 64.69 & 32.06 & 79.27 & 75.03 & 62.12 \\




Adamix &0.689\% &\textbf{35.14} & 76.99 & 89.01 & 67.10 & \textbf{38.41} & 79.49  & 75.09 & 64.33  \\

MoLoRA &0.746\% & 33.44 & 79.91 & 88.92 & 65.66 & 35.98 &   78.14 & 76.37 & 63.49  \\

SiRA & 0.746\% &33.98 & \textbf{81.26} & \textbf{89.04} & \textbf{68.10} & 37.71 & \textbf{82.17} & \textbf{75.50} & \textbf{65.13} \\

\bottomrule
\end{tabular}
\label{tab:multitask}
\end{table*}


\begin{table}[t]
\caption{Performance Comparison For Multilingual Tasks.}
\fontsize{9pt}{9pt}\selectfont
\begin{tabular}{l|c|c|c}
\toprule

       Approach   & $\delta$ params    & QA-in (9) & QA-cross (25)
\\ \midrule
LoRA & 0.043\% & 85.09 & 69.41  \\

Adamix & 0.689\% & 84.75 & 70.42  \\

MoLoRA &0.746\% &  85.14(WIP) & 69.70(WIP)  \\

SiRA & 0.746\% & \textbf{86.38} & \textbf{70.86}  \\

\bottomrule
\end{tabular}
\label{tab:multiligual}
\end{table}

\subsection{Evaluation Setup}

\paragraph{Baselines and Experiment Configs}

We specifically compare our model with the standard LoRA~\citep{hu2021lora}, Adamix~\citep{wang2022adamix} and MoLoRA~\citep{zadouri2023pushing}. 
Note that other adapter approaches are not commpared with as the SiRA approach is orthogonal and could be applied on top of them as well. 
We choose the PALM2-FLAN XXS~\citep{palm2} as the foundation model. We follow the default configurations in ~\citep{hu2021lora} to inject LoRA weights on the attention projections and set the intrinsic rank as $4$. We use $16$ experts by default across all baselines.
For training config and model selection, see Appendix~\ref{app:train}. 
 
\paragraph{Datasets and Metrics}

We evaluate on the following datasets:\footnote{Since our base model \cite{chung2022scaling} had been exposed to many public datasets during training, we choose dataset that are not consumed yet.}

\textbf{XTREME-UP} The XTREME-UP dataset \cite{ruder2023xtremeup} is a multilingual multitask dataset, with a focus on the scarce-data scenarios of underrepresented languages. In this work, we choose two of the underrepresented languages---Swahili(SW) and Bengali(BN)---and evaluated on several NLP tasks where these two languages have training and evaluation data. We follow \citet{ruder2023xtremeup} for each task's splits and evaluation metrics. 

\textbf{FinQA} FinQA \cite{chen-etal-2021-finqa} is a QA dataset in the financial domain. Complex reasoning capabilities are needed to correctly answer these questions. Note that the answers of the FinQA dataset are programs of a special arithmetic DSL. In this work we only evaluate on metrics based on surface form matching, \emph{i.e.}, exact match and F1 scores.

\textbf{ForumSum} ForumSum \cite{khalman-etal-2021-forumsum-multi} is a diverse and high quality conversation summarization dataset with human written summaries where the conversations are collected from a wide variety of internet forums. We report BLEURT~\citep{sellam2020bleurt}, ROUGEL, and F1 scores.

\subsection{Performance of SiRA}
\label{sec:quality}

We evaluate all the single tasks performance in Table~\ref{tab:single_task}.
Note that as FinQA is a hard task with financial reasoning, thus the exact match and f1 score is relatively low. 
We can notice that SiRA is outperforming all other baselines at most of the tasks, with less than $1\%$ extra parameters. 
Notably when compared to MoLoRA, SiRA achieves constantly better performance among all the tasks. This demonstrates that ``sparse" MoE is better than ``full" MoE. 
Adamix shows some small advantage on the Semantic Parsing task, but overall loses to SiRA across all other tasks. 

We also conducted experiments on two multitask settings on language swahili (SW) and bengali(BN), and two multiligual settings  for QA in languages task and QA across languages task. We reported numbers in Table~\ref{tab:multitask} and Table~\ref{tab:multiligual} respectively. The overall trend is similar to what we found in the single tasks. SiRA achieves the best average performance among all baselines.




\subsection{Ablation Study}

\paragraph{Computation Ablations}

\begin{table}[t]
\caption{Self ablations on the hyper-parameter topK(K) and expert capacity(C) on ForumSum.}
\centering
\fontsize{9pt}{9pt}\selectfont
\begin{tabular}{l|c|c|c|c|c}
\toprule
Configs & bleurt  & rougeL & f1
\\ \midrule                  

K=2 &96.87	& 34.51	&	24.73 \\
K=4 &96.60 &	34.66 &		25.34 \\
K=6 &96.75 &	34.73 &		24.55 \\
K=8 &96.76 &	\textbf{35.31} &		\textbf{25.64} \\
K=10 &\textbf{97.51} &	35.10	&	25.19 \\
K=12 &96.96 &	34.49	&	24.24 \\ \midrule

 C=2 &96.33	&34.15	&	24.13 \\
 C=4 &96.60&	34.66	&	25.34 \\
 C=6 &97.14	& \textbf{35.67}	& \textbf{25.83} \\
 C=8 &\textbf{97.31}&	34.97	&	25.24 \\
 C=10 &97.25&	34.75	&	25.57 \\
 C=12 & 96.50&	34.44	&	23.94 \\ 


\bottomrule
\end{tabular}
\label{tab:ablation_param}
\end{table}

We share the ablations on ForumSum in Table~\ref{tab:ablation_param}. 
We choose a simple config as base (k=4, C=4) and then change each of them while keeping the rest.
We first range the top $K$ from $2$ to $12$, with capacity $C=K$. And then we fix the $K=4$, and range the expert capacity from $2$ to $12$.  
An interesting finding is increasing the computation or the capacity will not always increase the scores and there is a ``comfortable zone'' which we need to find out with model tuning. 
This also justifies why the ``full'' MoE based approach is not as good as SiRA. SiRA provides more fine-grained control on the computation.  

\paragraph{Gating ablations}

\begin{table}[t]
\caption{Gating ablations on ForumSum.}
\centering
\fontsize{9pt}{9pt}\selectfont
\begin{tabular}{l|c|c|c}
\toprule
Approach & bleurt  & rougeL & f1
\\  \midrule
SiRA  &\textbf{97.14}	& \textbf{35.67}	& \textbf{25.83} \\

- aux loss&  96.37 &  35.09 & 25.11 \\
- Expert Dropout  & 97.09 & 34.73 & 24.55 \\
 + SMoE-Dropout &  96.30 & 34.24 & 24.32 \\
\bottomrule
\end{tabular}
\label{tab:gating_ablation}
\end{table}

We also provide the ablations on the gating in Table~\ref{tab:gating_ablation}. Specifically we compare SiRA with 3 cases: 1/ removing the aux loss, 2/ removing the gate dropout, and 3/ using a static routing based dropout SMoE-Dropout~\citep{chen2023sparse} instead. 
Results suggested that the learned gating is still better than a fixed one, and both the gate dropout and aux loss help the performance of SiRA. 

\paragraph{What the gate learns}
We use the Swahili multitask experiment to study what the gate is learning. 
We measure the average entropy of each gate weight distribution before TopK is applied. The average entropy for the QA (in language) task decreases from 1.6 to 1.13 nats during training. This indicates that the model learns to give certain gates more weight as it trains.

We also measure the average correlation coefficients between each task index and each gate index similar to \citep{mod_squad}. We convert the task index to a one hot encoding for this. 
At the end of training, the average correlation was about .025, which is not significant. The correlation between gates and languages in the multilingual experiment is not significant either. 
This suggests that our gating mechanism does not learn to route different tasks to different gates.



\section{To-Do List}
This manuscript is currently under active development. Our upcoming endeavors include getting more results and analysis, and improving the writings.
We warmly welcome suggestions, insights, and constructive criticism from the research community. Should you have any feedback or ideas that could potentially enhance the quality and impact of our work, please do not hesitate to contact the lead author. Your input is invaluable to us, and we are eager to integrate diverse perspectives to refine and advance our research.

\section{Conclusion}

This paper introduced SiRA, a Sparse Mixture of Expert variant of LoRA. 
SiRA enforces the top $k$ experts routing with capacity constraint for each experts. We also devise a novel expert dropout mechanism on top of the auxiliary loss to reduce its over-fitting issue.  
We conducted extensive experiments to verify the performance of SiRA, which achieves better performance than the LoRA and its MoE variants across different single tasks and multitask settings.

\section{Limitation}
SiRA is taking extra serving overhead for serving with extra parameters on experts and the gating, compared to LoRA or Adamix. 
How to minimize the serving overhead is a challenge problem which we hope to address our future works.

\section*{Acknowledgements}

We would like to acknowledge Abhanshu Sharma, Hassan Mansoor, Qifei Wang, Victor Cărbune etc for their valuable inputs.

\bibliography{anthology,custom}

\section{Appendix}
\label{sec:appendix}

\subsection{Effect of LoRA rank}
\label{app:comp}

\begin{figure}[ht]
\includegraphics[width=0.98\columnwidth]{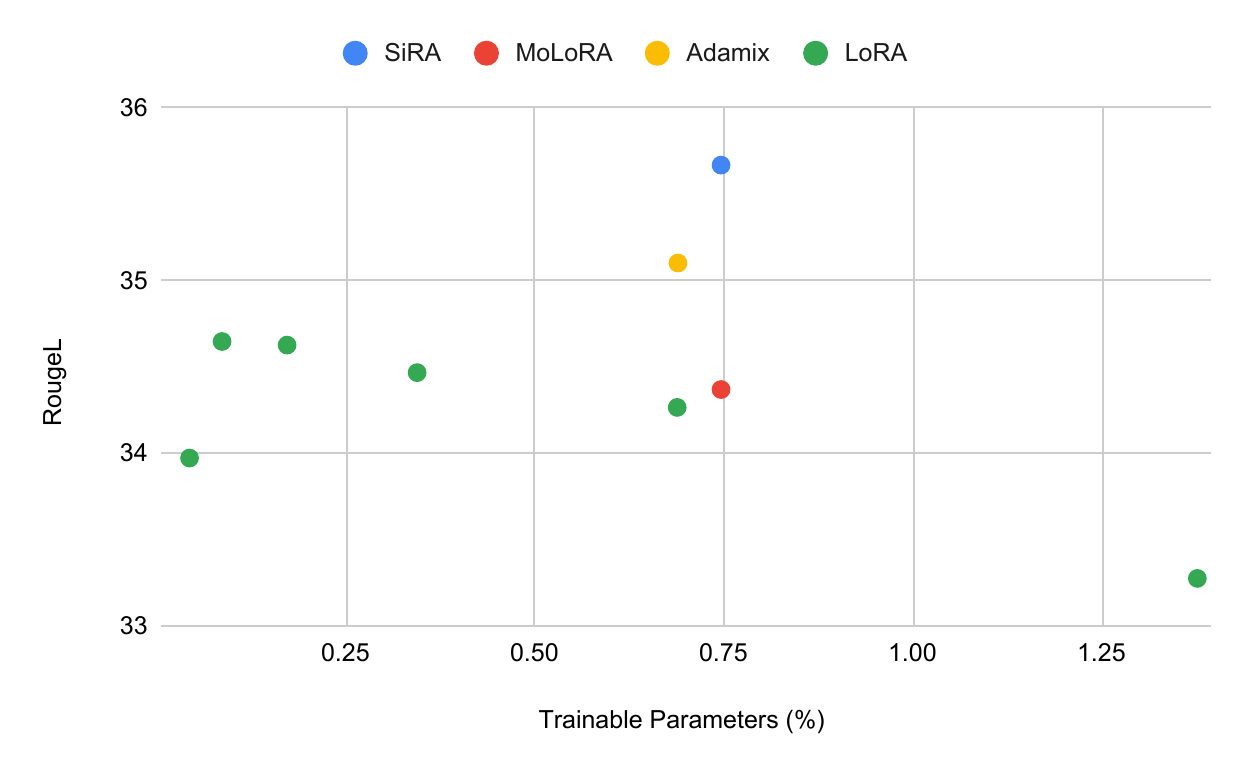}
\centering
\caption{SiRA vs LoRA on ForumSum Task. We increase the rank of LoRA (rank=4, 8, 16, 32, 64, 128) and report the RougeL as a metrics. Notably increasing the rank does does not help the performance. SiRA (rank=4) can achieve higher quality by leveraging the sparse mixture of experts.
}
\label{fig:comp}
\end{figure}

We investigate the effect of LoRA rank in Figure~\ref{fig:comp}.

\subsection{Training and Model selection}
\label{app:train}

During supervised finetuning, SFT, we use $8$ Tensor Processing Units (TPU) V3 chips for fine-tuning. The batch size is $64$, and the maximum training step is $30000$. We use the Adafactor optimizer~\citep{shazeer2018adafactor} with a learning rate of $0.0005$. Both the input and output sequence lengths are set to match the dataset requirements. 
The training dropout rate is $0.05$. The expert dropout rate is set to $0.5$. 
We did hyper-parameters search to find the best model configurations.
We decode on the validation sets of each task every $100$ steps. And we report test results from the best checkpoints according to the validation scores. For multitask results, the checkpoint is picked on the average each tasks metrics.
For the reported numbers in section~\ref{sec:quality}, we use topk $K=4$ as default. Yet we found $K=8$ is better for BN multitask and QA (in-lang) multilingual setting, and $K=12$ better for QA (cross-lang) experiments. 

\end{document}